\documentclass{article}

\usepackage{microtype}
\usepackage{graphicx}
\usepackage{subfigure}
\usepackage{booktabs} 
\usepackage{soul}
\usepackage{color}
\usepackage[dvipsnames]{xcolor}
\usepackage{bm}
\usepackage{amsmath}
\usepackage{multicol}
\usepackage{multirow}
\usepackage[titletoc]{appendix}
\usepackage{multibib}
\newcites{sec}{References}

\newcites{SM}{References}

\usepackage{hyperref}

\usepackage[accepted]{icml2020}

\icmltitlerunning{T-GD: Transferable GAN-generated Images Detection Framework}

\newcommand{\sysname}{T-GD}

\begin{document}
\twocolumn[
\icmltitle{T-GD: Transferable GAN-generated Images Detection Framework}



\begin{icmlauthorlist}
\icmlauthor{Hyeonseong Jeon}{ai}
\icmlauthor{Youngoh Bang}{ai}
\icmlauthor{Junyaup Kim}{sw}
\icmlauthor{Simon S. Woo}{sw,ds}
\end{icmlauthorlist}

\icmlaffiliation{ai}{Department of Artificial Intelligence, Sungkyunkwan University, Suwon, S. Korea}
\icmlaffiliation{sw}{Computer Science and Engineering Department, Sungkyunkwan University, Suwon, S. Korea}
\icmlaffiliation{ds}{Department of Applied Data Science, Sungkyunkwan University, Suwon, S. Korea}

\icmlcorrespondingauthor{Simon S. Woo}{swoo@g.skku.edu}

\icmlkeywords{Machine Learning, ICML}

\vskip 0.3in
]



\printAffiliationsAndNotice{}  

\begin{abstract}
Recent advancements in Generative Adversarial Networks (GANs) enable the generation of highly realistic images, raising concerns about their misuse for malicious purposes. Detecting these GAN-generated images (GAN-images) becomes increasingly challenging due to the significant reduction of underlying artifacts and specific patterns. The absence of such traces can hinder detection algorithms from identifying GAN-images and transferring knowledge to identify other types of GAN-images as well. In this work, we present the Transferable GAN-images Detection framework~(\sysname), a robust transferable framework for an effective detection of GAN-images. \sysname~is composed of a teacher and a student model that can iteratively teach and evaluate each other to improve the detection performance. First, we train the teacher model on the source dataset and use it as a starting point for learning the target dataset. To train the student model, we inject noise by mixing up the source and target datasets, while constraining the weight variation to preserve the starting point. Our approach is a self-training method, but distinguishes itself from prior approaches by focusing on improving the transferability of GAN-image detection.~\sysname~achieves high performance on the source dataset by overcoming \textit{catastrophic forgetting} and effectively detecting state-of-the-art GAN-images with only a small volume of data without any metadata information.
\end{abstract} 
\section{Introduction}

\begin{figure}[t]
\begin{center}
   \includegraphics[width=1.0\linewidth]{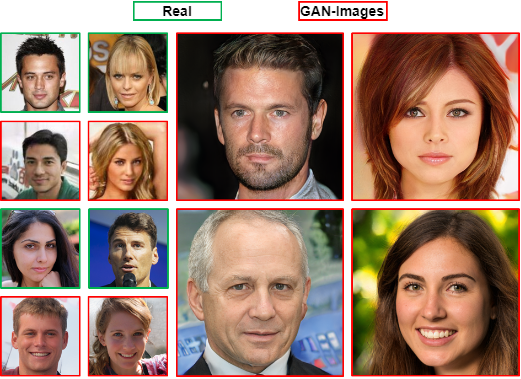}
\end{center}
   \caption{Sample data used for our experiment. Images inside the green border are real images, while those inside the red border are GAN-images. The left-hand side images in row order are from CelebA, StarGAN, FFHQ, and StyleGAN. The right-hand side images in row order are from PGGAN and StyleGAN2.}
\label{fig:selfattention}
\end{figure}

Recent advancements in Generative Adversarial Networks (GANs)~\cite{choi2019stargan, zakharov2019few, karras2019analyzing, shaham2019singan} enable the generation of realistic images, which has now become feasible through few-shot or single-shot learning. Some GANs manage to further reduce visible artifacts and patterns, such as blurred object shape, checkerboard artifacts, semantically strange objects, and unnatural backgrounds. For these reasons, even high-resolution images produced by the latest GANs are hardly distinguishable from real images or by human inspection. 

A typical way of detecting GAN-images is to train Convolutional Neural Networks (CNNs) and a binary classifier with a large number of images generated from GANs. Some researchers~\cite{marra2019gans, zhang2019detecting, yu2019attributing} have shown that the detection performance can be improved by analyzing artifacts and patterns in GAN-images. Many of the existing methods has achieved high performance in detecting GAN-images when the model tests on the same dataset used during the training phase~\cite{tariq2019gan, tariq2018detecting, jeon2019faketalkerdetect}. Moreover, this binary classifier can be realized by the use of existing and well-structured CNN architectures~\cite{tan2019efficientnet, jeon2020fdftnet}.

\begin{figure*}[t]
    \centering
      \includegraphics[width=0.9\linewidth]{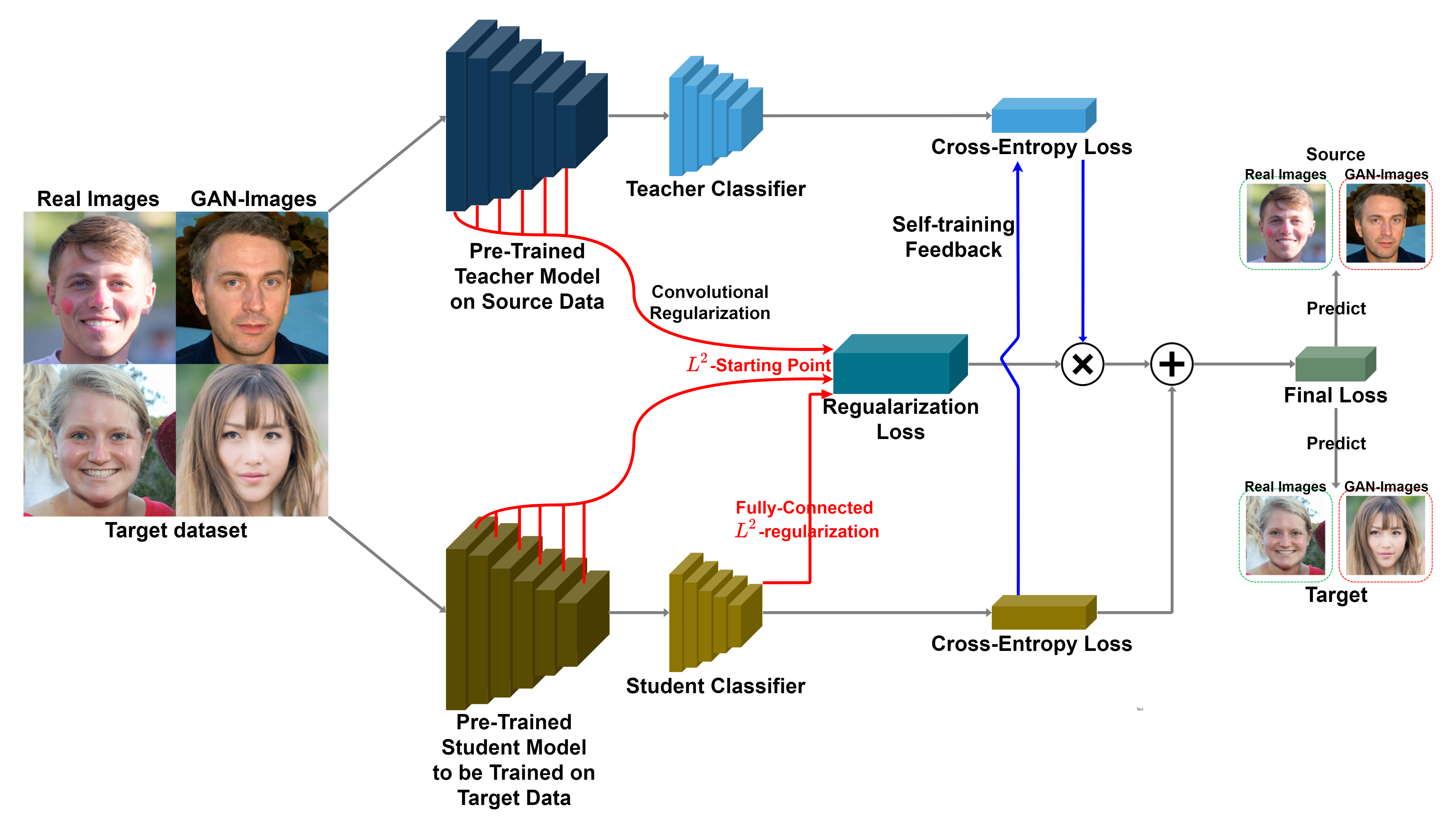}
    \caption{Overview of our T-GD network. For efficient transfer learning, our network uses \textit{$L^2$-Starting Point} (red) and \textit{Self-training} (blue).}
    \label{fig:overview_tgd}
\end{figure*}

However, above methods are ineffective for improving transfer learning performance. That is, when the CNN classifier is trained on one dataset, it shows poor performance on other datasets. ForensicTransfer~\cite{cozzolino2018forensictransfer} introduced an autoencoder for the GAN-image detection. They apply autoencoder and detect GAN-images through reconstruction error. This learning method has advantages regarding lower data usage, when the model is well trained. Although the ForensicTransfer showed promise for model transferability, its performance remains mediocre. In previous research, either artifacts, patterns, or augmentations were utilized individually for successful transfer learning, yet it is possible to combine them to transfer knowledge of GAN-image detection.

In this paper, our objective is to maintain a high detection performance during transfer learning on the source and target datasets, without suffering \textit{catastrophic forgetting}. While many studies on transfer learning have already shown impressive performance, they have not applied for GAN-image detection. Therefore, in this work, we propose a novel regularization method with self-training for transfer learning by combining and transforming regularization, augmentation, self-training, and learning strategies to improve transferability of GAN-image detection. In particular, our approach is inspired by starting point as the reference (SPAR), $L^2$-$SP$~\cite{li2018explicit}, which regularizes the weight variation of a target model by referring to the weights pre-trained on the source dataset. The limitation of the latter method is that it cannot provide an optimal solution for transfer learning. As the regularization strength changes, the control of the regularization can easily be lost. Our approach overcomes this issue through self-training, where the teacher model automatically helps control the strength of regularization when the student model learns from the target dataset.

In addition, we introduce a novel augmentation method to solve the over-fitting problem by transforming Cutmix~\cite{yun2019cutmix}, which randomly mixes up a rectangle patch of training images. Note that the original Cutmix mixes up inter-class images; our experiment showed that this renders the learning process highly unstable for a binary classifier, and thus we transform the inter-class Cutmix to an intra-class Cutmix to increase the stability of the learning process. Also, our method combines Gaussian blur~\cite{xuan2019generalization} and Joint Photograph Experts Group (JPEG) compression~\cite{wang2019cnn}, which were previously studied for GAN-image detection.

For transfer learning, we apply learning strategies, such as Weight Standardization~\cite{qiao2019weight} (WS), Group Normalization~\cite{wu2018group} (GN), and the tuning of learning rate and momentum rate~\cite{li2020rethinking}. WS and GN achieved comparable performance on image classification, not depending on batch size statistics. Also, we implement transfer learning using low learning and momentum rates for stochastic gradient descent (SGD) inspired by~\cite{li2020rethinking}. These have been experimented on object detection transfer, but we demonstrate that these strategies also work well for transfer learning across different domains. Finally, we integrate these approaches into one framework, the Transferable GAN-generated image Detection framework (\sysname). Compared to general methods of transfer learning and recent GAN-image detection, we show that~\sysname~is equipped with robust transferability and achieves high performance.


\section{Related Work}
\textbf{GAN-image detection.\hspace{1mm}} Widely used approaches for the detection of GAN-images include the addition of a learning method, the transformation of GAN-images, the application of data augmentation, and the use of metadata. Another technique, based on multi-task incremental learning~\cite{marra2019incremental}, shows great promise for transferability within different types of GAN-images, changing the existing learning method (autoencoder) and loss function (incremental learning). However, a lingering issue is the need for a large amount of data; our work directly alleviates this problem, maintaining the model performance with a smaller amount of data. \cite{nataraj2019detecting} proposed using a combination of co-occurrence matrices and JPEG compression to transform GAN-images for data augmentation. Co-occurrence matrices are extracted from three color channels in the pixel domain following JPEG compression and are used to train the CNN. In their approach, the JPEG compression contributed to the improvement of performance, but the transformation of the input data into co-occurrence matrices caused over-fitting and reduced generalizability.

Some methods identify a unique artifact spectrum caused by the up-sampling component~\cite{zhang2019detecting}, while others use the photo-response non-uniformity (PRNU) pattern as the input of CNN classifiers~\cite{marra2019gans}. Augmentation techniques requiring domain knowledge of GAN-image detection, such as Gaussian blur and Gaussian noise~\cite{xuan2019generalization}, were also studied and the combination of Gaussian blur and JPEG compression was shown to achieve high performance~\cite{wang2019cnn}. However, employing a single data augmentation method achieved limited transfer learning performance. Therefore, we use the combination of JPEG compression and Gaussian blur to achieve better transfer learning results. Similar to prior digital fingerprint techniques, GAN fingerprints (i.e., image and model fingerprints) are used to differentiate real and GAN-images using metadata~\cite{yu2019attributing}. They assume \textit{white-box attack scenarios}, where detectors possess knowledge of the data and the model (metadata information) of attackers. Our approach differs from theirs, where ours is constrained to \textit{a non-adaptive black-box approach scenario}; detectors only possess knowledge of the training data.

\textbf{Transfer learning.\hspace{1mm}} Several transfer learning methods have been explored in terms of meta-learning, self-supervised learning, domain adaptation~\cite{zamir2018taskonomy}, knowledge distillation, and continual learning. In particular, starting point as the reference (SPAR)~\cite{li2018explicit} and Deep Learning Transfer using Feature Map with Attention (DELTA)~\cite{li2019delta} use $L^2$ regularization as a starting point to maintain the source dataset as the inductive learning method during domain adaptation. The difference from ours is that we use self-training to control the regularization effect, which has the advantage of preventing either excessive or minor regularization. 

\cite{yim2017gift} proposed two additional layers to calculate the flow of solution procedure matrix for knowledge distillation, but our T-GD shows robust performance in transfer learning without these layers and expensive computation. Learning with continual tasks~\cite{zenke2017continual}, where the node weight is regularized based on the importance of previous tasks, is similar to our method. However, we provide the following differences: First, the source and target datasets differ in size (see Table~\ref{table:dataset}), resulting in relatively low computational cost in $L^2$-$SP$. Second, in our task, achieving generalizability using a small amount of target data is as essential as the prevention of forgetting; to address this trade-off, we chose $L^2$-$SP$ using all weights. Third, unlike continual learning on each independent task, our T-GD focuses on transfer learning within the GAN-image domains.

\textbf{Self-training.\hspace{1mm}} Self-training methods~\cite{yalniz2019billion, xie2019self} were used to increase the state-of-the-art top-1 accuracy of ImageNet~\cite{ILSVRC15}. The difference from our work is that our objective is to increase transferability and not the single model performance. We use a teacher-student structure, and inject noise to the student model and the input data to prevent over-fitting by effective techniques, such as dropout~\cite{srivastava2014dropout}, stochastic depth~\cite{huang2016deep}, and intra-class Cutmix (data augmentation)). Also, self-training was used for domain adaptation by~\cite{roychowdhury2019automatic}. Their self-training model is video-specific, applying a teacher model to the target domain, which is different from our image classification task.

\section{Our Method}

The first step of~\sysname~is to train the pre-trained models, namely the binary classifiers predicting whether an image is GAN-generated or not.

\textbf{CNN binary classifier.\hspace{1mm}} We chose CNN binary classifiers as classifiers for the source dataset. This choice has three advantages: (1) it is easy to reuse pre-trained models, (2) many pre-studied CNN architectures can be utilized, and (3) it shows a more stable performance in binary classification than other methods such as autoencoders. We pre-train the CNN binary classifier on the source dataset and transfer (fine-tune) this pre-trained model to the target dataset. For instance, EfficientNet (the CNN classifier) is trained on the PGGAN-dataset (the source) and fine-tuned on the StyleGAN-dataset (the target). 

\textbf{EfficientNet.\hspace{1mm}} We implemented EfficientNet-B0~\cite{tan2019efficientnet} and used it as the CNN classifier. Although EfficientNet-B0 has the lowest number of parameters (about four million) among the EfficientNets, it performs well in GAN-image detection in our experiment, compared to Inception-V3~\cite{szegedy2017inception} and Xception~\cite{chollet2017xception}. Another change we make to the model is the use of WS~\cite{qiao2019weight} and GN~\cite{wu2018group}, instead of batch normalization (BN)~\cite{ioffe2015batch}, due to their superior efficiency regarding transfer learning to that of batch statistics.

\textbf{ResNext.\hspace{1mm}} We implemented ResNext32$\times$4d~\cite{xie2017aggregated} and used it as the CNN classifier, where ResNext32$\times$4d has more parameters (about twenty million) than EfficientNet-B0. We also replace BN with WS and GN.
 

\subsection{\bm{$L^2$}-\bm{$SP$}}
The next step is transfer learning. The weight of the pre-trained model from the source dataset is used as the SPAR. In particular, we use $L^2$-$SP$ for transfer learning. Regularization can lead to a better optimization by preventing over-fitting when learning from scratch; $L^2$-$SP$ differs in that the starting point from a well pre-trained source dataset guides the learning process by referring to the information of the pre-trained source dataset. This method does not require freezing the weights of the pre-trained model nor using weight decay. Our method regularizes convolution layers and fully-connected (FC) layers independently.

\textbf{General form of regularization.\hspace{1mm}} Let $w$ be the weight parameters, and $J(\hat{y_i}, y_i)$ be the loss function of the neural networks, where $\hat{y_i}$ is the $i^{th}$ score predicted by the models and $y_i$ is the $i^{th}$ label. And $\Omega(w)$ is the $p$-norm function of the weight $w$ as a general form of regularization loss, $f_w(x_i)$ is the neural network function with the $i^{th}$ data $x_i$, and $n$ is the dataset size.
Equation~\ref{eq:general} indicates the general form of the loss function with a weight regularization component:
\begin{equation}
    \begin{split}
        \label{eq:general}
        \min_w&\frac{1}{n}\sum^{n}_{i=1}J(\hat{y_i}, y_i)+\lambda\cdot\Omega(w), \\
        \hat{y_i} &= f_w(x_i),
    \end{split}
\end{equation}
where $\lambda$ balances the regularization and the loss function, $J$ is the cross-entropy function, and $\Omega$ is the $L^1$ or $L^2$-norm of the parameter $w$.

\textbf{\bm{$L^2$} regularization.\hspace{1mm}} $L^2$ regularization is used in transfer learning to avoid over-fitting and to overcome the forgetting of the learned information or \textit{catastrophic forgetting}.
\begin{equation}
    \label{eq:l2reg}
    \Omega_{l2}(w)=\|w\|^2_2.
\end{equation}
Equation~\ref{eq:l2reg} is the $\Omega$ function, namely the $L^2$-norm of $w$.
\begin{equation}
    \label{eq:clsreg}
    \min_w\frac{1}{n}\sum^{n}_{i=1}J(\hat{y_i}, y_i)+\beta\cdot\Omega_{l2}(w_{fc}).
\end{equation}
In Eq.~\ref{eq:clsreg}, the first term is the same as in Eq.~\ref{eq:general}, representing the cross-entropy loss function. The second term is the $\Omega_{l2}$ function or the $L^2$ regularization term (Eq.~\ref{eq:l2reg}) of $w_{fc}$, the weights of the FC layers, scaled by $\beta$, which is equivalent to $\lambda$ in Eq.~\ref{eq:general}. Note that the $L^2$ regularization is applied solely to the FC layers since over-fitting and forgetting are delayed, but not completely prevented in the course of learning.
 
\textbf{\bm{$L^2$}-\bm{$SP$}.\hspace{1mm}} Let $w^\prime$ be the pre-trained weights from the source dataset, as shown in Section 3.1, serving as the starting point (SP) as the reference, as well as a regularization point that provides guidance for transfer learning when fine-tuning. Using L2-norm, we define $L^2$-$SP$ as follows:
\begin{equation}
    \label{eq:l2sp}
    \Omega_{sp}(w,w^\prime)=\|w_{conv}-w^\prime_{conv}\|^2_2,
\end{equation}
where $w_{conv}$ denotes the weights of the convolution layers, excluding those of the FC layers. Equation~\ref{eq:l2sp} indicates that $L^2$-$SP$ is a one-to-one mapping between the convolution layers of the source and target datasets, e.g., the PGGAN-classifier (the source) to the StyleGAN-classifier (the target).

\textbf{Loss function.\hspace{1mm}} We combine Eq.~\ref{eq:l2sp}, sharing the architecture of the source and target models, with the second term of Eq.~\ref{eq:clsreg}, accounting for the FC layer (final layer) as follows:
\begin{equation}
    \label{eq:loss}
    \begin{split}
    \min_w\frac{1}{n}\sum^{n}_{i=1}J(\hat{y_i},y_i)+\alpha\cdot\Omega_{sp}(w,w^\prime)+\beta\cdot\Omega_{l2}(w_{fc}), \\
    J(\hat{y_i},y_i)=-y_i\log(\hat{y_i})-(1-y_i)\log(1-\hat{y_i}),
    \end{split}
\end{equation}
where $J$ is the negative log-likelihood loss function, and $\alpha$ and $\beta$ are tunable hyperparameters of which~\cite{li2019delta} use values in the range from 0.1 to 0.01. The difference from $L^2$-$SP$ is that we transform $\alpha$ and $\beta$ into $\gamma$, a parameter which adjusts itself according to the learning situation. More details about the transformed parameters are provided in Section 3.3.

\begin{figure}[t]
\begin{center}
   \includegraphics[width=1\linewidth]{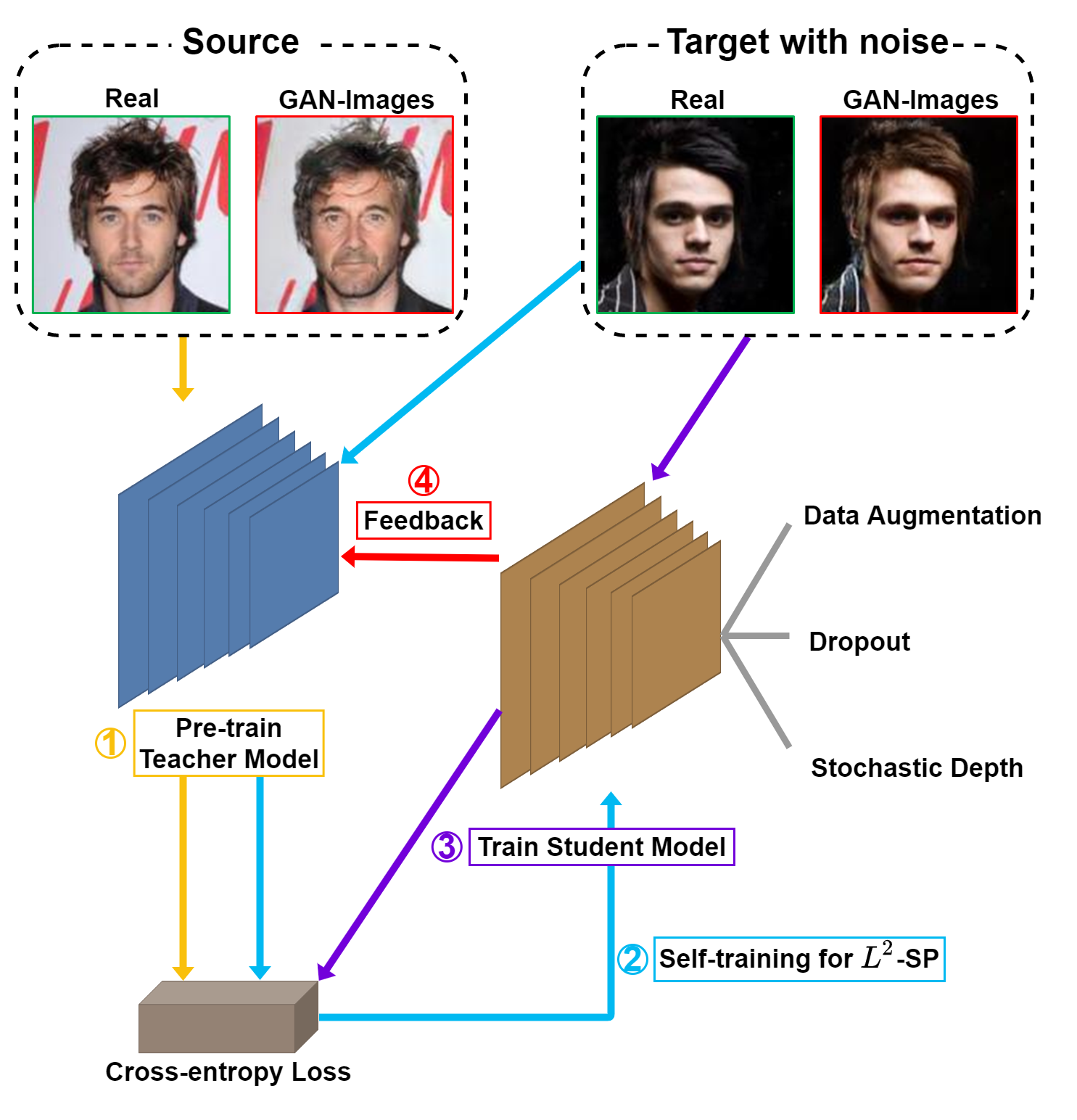}
\end{center}
   \caption{Overview of our self-training method. Note that the number of processes shown in this figure equals to that of the ordered processes demonstrated in Alg.~\ref{alg:selftraining}.}
\label{fig:selftraining}
\end{figure}

\subsection{Self-training for \bm{$L^2$}-\bm{$SP$}}
We transform the transfer learning framework into a self-training framework. In other words, the source/target model is changed into a teacher/student model. In addition to the role of a typical source model, which serves as SPAR and a regularizer to guide the learning process, the teacher model has the role of adjusting the parameters based on the learned target dataset (training loss). That is, the teacher model directly controls $\alpha$ and $\beta$, as shown in Eq.~\ref{eq:loss}.

Let $\{(x_1, y_1),(x_2,y_2),...,(x_n,y_n)\}$ be the labeled source dataset and $\{(\tilde{x}_1,\tilde{y}_1),(\tilde{x}_2, \tilde{y}_2),...,(\tilde{x}_m,\tilde{y}_m)\}$ the labeled target dataset. In a typical self-training process, unlabeled data is used to increase the generalizability for single dataset performance, e.g., ImageNet, by learning from extra training data. However, we assume the usage of additional data is highly limited. Hence, we only use labeled data for transfer learning. 

\begin{equation}
    \label{eq:self}
    \begin{split}
    J(\hat{y_i},\tilde{y}_i)&=-\tilde{y}_i\log(\hat{y_i})-(1-\tilde{y}_i)\log(1-\hat{y_i}), \\
    \hat{y_i} &= f_{w^\prime}^{noised}(\tilde{x}_i^{noised}). \\
    \end{split}
\end{equation}

In Eq.~\ref{eq:self}, $w^\prime$ denotes the weights of the teacher model, and $f_{w^\prime}$ denotes the pre-trained models from the source dataset. $J(\hat{y_i},\tilde{y}_i)$ denotes the binary cross-entropy loss, but the input data, $\tilde{x}_i^{noised}$, is from the target dataset with noise injection.
\begin{equation}
    \label{eq:gamma}
    \begin{split}
        \gamma&:=s\sigma \left(-\frac{1}{m}\sum_{i=1}^{m}J(\hat{y_i},\tilde{y}_i)\right), \\
        \sigma&(x)=1/(1+e^{-x}).
    \end{split}
\end{equation}
In Eq.~\ref{eq:gamma}, $s$ is a hyperparameter taking values from 0.1 to 2.0, and the $\gamma$ score is obtained through the sigmoid function $\sigma$ to help stabilize training, whose input is the negative mean loss function described in Eq.~\ref{eq:self}. In the transfer phase, we use the same noised target data for both the teacher and student models. The teacher is evaluated on the data (Eq.~\ref{eq:self}), and the negative value of the result is taken and transformed by the sigmoid function $\gamma$ in Eq.~\ref{eq:gamma}, where $\gamma$ regulates the intensities of both $L^2$-$SP$ (Eq.~\ref{eq:l2sp}) and the L2-norm of FC layer (Eq.~\ref{eq:clsreg}). An analysis of $\gamma$ and the error amplification of self-training is presented in Supp. A.

\textbf{Final loss function.\hspace{1mm}}  We replace $\alpha$ and $\beta$ with $\gamma$ in Eq.~\ref{eq:loss} to act as a changeable balancing parameter for regularization as follows:
\begin{equation}
    \label{eq:final_loss}
    \min_w\frac{1}{n}\sum^{n}_{i=1}J(\hat{y_i},y_i)+\gamma\cdot\Omega_{sp}(w,w^\prime)+\gamma\cdot\Omega_{l2}(w_{fc}).
\end{equation}
The final loss function, as shown in Eq.~\ref{eq:final_loss}, is composed of a cross-entropy term and an $L^2$-$SP$ term for the self-training of the student model. Figure~\ref{fig:overview_tgd} shows an overview of this entire pipeline, and Supp. Alg.~\ref{alg:selftraining} presents the detailed algorithm. 

\begin{figure}[t!]
\begin{center}
   \includegraphics[width=1.0\linewidth]{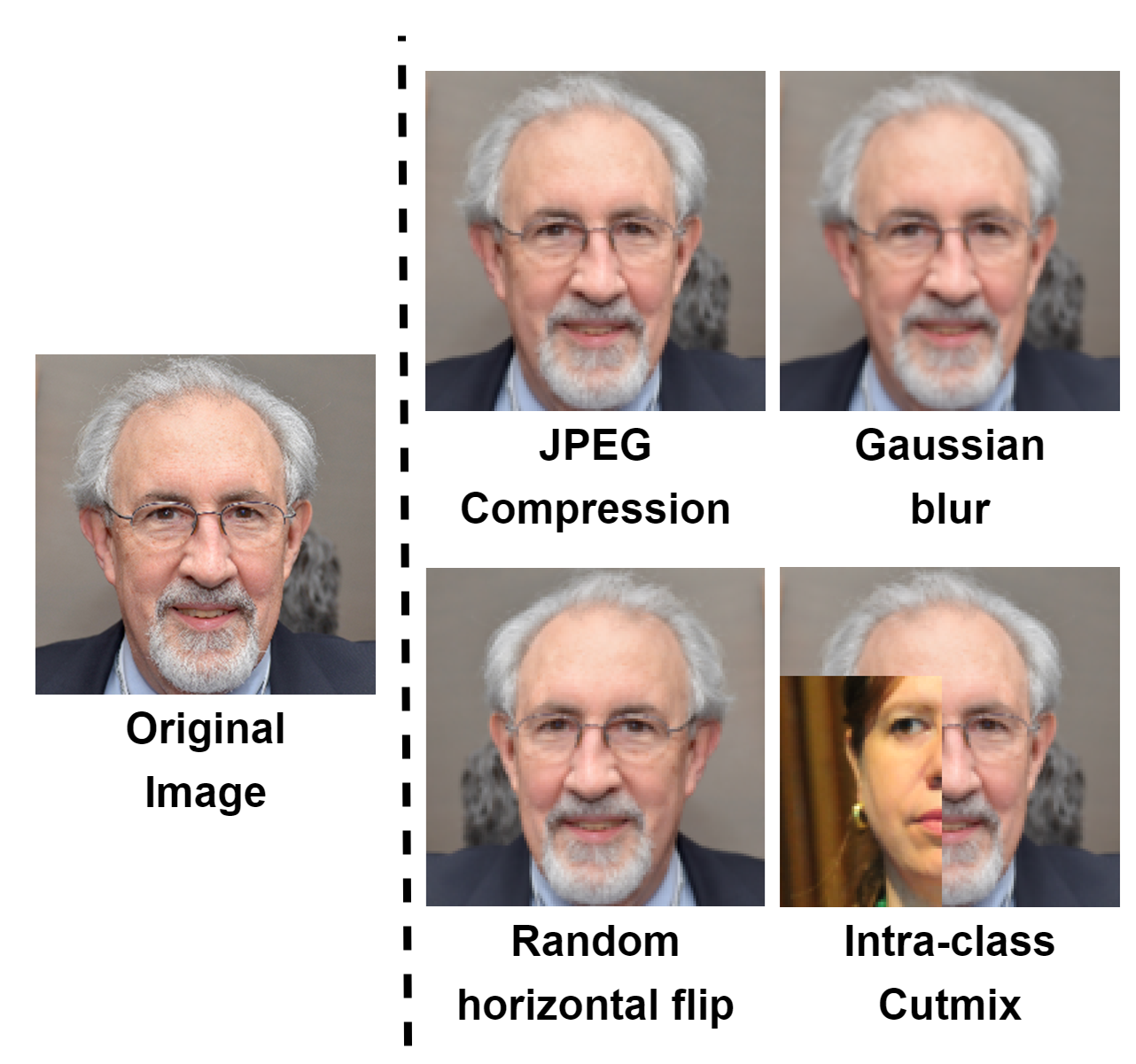}
\end{center}
   \caption{Examples of different noisy data augmentation techniques applied to the original input image.}
\label{fig:noisy_student}
\end{figure}

\begin{table*}[t]
\centering
\renewcommand{\arraystretch}{1.2}
    \resizebox{\linewidth}{!}{%
    \begin{tabular}{l|l|cccc|cccc}
    \toprule
      \multirow{2}{*}{Method} & \multirow{1}{*}{Category} & \multicolumn{4}{c|}{Zero-shot (Pre-trained model)} & \multicolumn{4}{c}{Transfer Learning} \\ \cline{2-10} & Dataset & PGGAN & StarGAN & StyleGAN & StyleGAN2  & PGGAN & StarGAN & StyleGAN & StyleGAN2  \\
    \midrule
    GeneralTransfer             
    & PGGAN & 99.91\% & 56.81\% & 49.47\% & 49.32\% &                           \underline{99.86\%} & 87.06\% & 54.17\% & 54.18\% \\
    \textit{EfficientNet-B0}     
    & StarGAN & 66.47\% & 99.88\% & 52.01\% & 52.10\% 
    & 95.90\% & \underline{89.87\%} & 99.03\% & 99.04\% \\                      (Base model)
    & StyleGAN & 49.80\% & 50.04\% & 99.96\% & 99.97\% 
    & 66.89\% & 51.12\% & \underline{99.94\%} & 99.95\%   \\
    & StyleGAN2 & 45.23\% & 49.00\% & 99.99\% & 99.99\% 
    & 91.33\% & 88.16\% & 45.26\% & \underline{47.37\%} \\
    \hline
    ForensicTransfer$\dagger$    
    & PGGAN & 97.15\% & 50.27\% & 53.57\% & 53.27\% 
    &  \underline{69.35\%} & 72.40\% & 76.50\% & 76.50\% \\
    & StarGAN & 47.09\% & 85.34\% & 49.51\% & 49.48\% 
    & 90.14\% & \underline{51.32\%} & 53.14\% & 53.14\% \\
    & StyleGAN & 49.23\% & 49.66\% & 99.12\% & 99.97\% 
    & 76.57\% & 58.93\% & \underline{65.83\%} & 65.85\% \\
    & StyleGAN2 & 49.22\% & 49.66\% & 99.12\% & 99.12\% 
    & 76.58\% & 58.94\% & 65.84\% & \underline{65.84\%} \\
    \hline
    \textbf{\sysname}            
    & PGGAN & 99.91\% & 56.81\% & 49.47\% & 49.32\% &                           \textbf{\underline{95.87\%}} & 91.61\% & \textbf{98.12}\% & \textbf{98.13\%}\\
    \textit{EfficientNet-B0}     
    & StarGAN & 66.47\% & 99.88\% & 52.01\% & 52.10\% 
    & 94.94\% & \textbf{\underline{97.32\%}} & 97.29\% & 93.34\%\\
    (Base model)                 
    & StyleGAN  & 49.80\% & 50.04\% & 99.96\% & 99.97\% 
    & 84.92\%   & 90.00\% &  \textbf{\underline{97.83\%}} & 97.71\% \\
    & StyleGAN2 & 45.23\% & 49.00\% & 99.99\% & 99.99\% 
    & 84.91\% & 90.01\% & 97.83\% &  \textbf{\underline{97.71\%}} \\
    \hline
    \textbf{\sysname}            
    & PGGAN & 99.81\% & 61.25\% & 49.76\% & 49.91\% 
    & \underline{94.91\%} & 93.21\% & 87.37\% & 87.58\% \\
    \textit{ResNext32$\times$4d} 
    & StarGAN & 41.43\% & 99.78\% & 48.37\% & 48.50\% 
    & 98.88\% & \underline{96.15\%} & 91.48\% & 91.26\% \\
    (Base model)                 
    & StyleGAN  & 41.05\% & 49.16\% & 99.99\% & 99.99\% 
    & 85.93\% & 79.69\% &  \underline{94.31\%} & 94.31\%  \\
    & StyleGAN2 & 38.90\% & 50.31\% & 99.90\% & 99.88\% 
    & 87.20\% & 80.19\% & \textbf{98.39\%} & \underline{95.38\%}  \\
    \bottomrule
    \end{tabular}}
    \caption{Performance results. ``$\dagger$'' indicates our implementation. All 4 GAN datasets are evaluated with 4 models as well as our models. The Dataset column indicates pre-trained model from a source dataset, and the Dataset row indicates the target test set for transfer learning. The evaluation metric is AUROC (\%). The underlined results are the source dataset performance after transfer learning. The best results are highlighted in bold. The Zero-shot category represents the performance of a pre-trained model without any additional training and the Transfer learning category represents each pre-trained model transferred from the source to target dataset.}
\label{table:performance}
\end{table*}

\textbf{Augmentation and noised model.\hspace{1mm}} Noise is injected to both the data and the model. For data augmentation, we use JPEG compression~\cite{wang2019cnn}, and Gaussian blur~\cite{xuan2019generalization}, random horizontal flip, and a transformed version of Cutmix~\cite{yun2019cutmix}, called intra-class Cutmix. We apply dropout~\cite{srivastava2014dropout} to the FC layer at a stronger rate than for the pre-trained model. In addition, we apply stochastic depth~\cite{huang2016deep}, randomly dropping the paths of residual layers, also at a stronger rate than for the pre-trained model.

\textbf{Intra-class Cutmix algorithm.\hspace{1mm}} The pseudo-code of the intra-class Cutmix algorithm is shown in Supp. Algorithm~\ref{alg:cutmix}. First, the mini-batch data is shuffled and the index at which the target label $Y_m$ equals to the shuffled target label $Y^\prime_m$, i.e., real to real and GAN-image to GAN-image, is denoted as same\_index in Algorithm~\ref{alg:cutmix}. If a random variable $\rho$ drawn from a uniform distribution between $0$ and $1$ is greater than the fixed Cutmix parameter (0.2 when pre-training, and 0.5 when transfer learning), then the input $X_m$ and the shuffled input $X^\prime_m$ are mixed by replacing a randomly cropped region of the input $X_m$ to the region of the shuffled input $X^\prime_m$. Cutmix~\cite{yun2019cutmix} mixes the target label $Y_m$ through interpolation. Intra-class Cutmix does not mix the label, because the input data $X_m$ still belongs to the same class following replacement.
\begin{table}[t]
\centering
    \begin{small}
                \begin{tabular}{l|ccc|c}
                \toprule
                 & \multicolumn{3}{|c|}{Source Data} & Target Data \\
                \midrule
                Dataset & Train & Validation & Test & Transfer \\
                \midrule
                PGGAN   & 64,202    & 16,051 & 18,799 & 2,000\\
                StarGAN & 137,239 & 15,260 & 50,000 & 2,000\\
                StyleGAN    & 33,739  & 3,900 & 30,000 & 2,000\\
                StyleGAN2   & 42,356  & 3,900 & 30,000 & 2,000\\
                \bottomrule
                \end{tabular}
    \end{small}
    \caption{GAN-generated datasets used in our experiment, where train, validation, test, as well as transfer dataset are shown. We only use 2,000 images for transfer learning.}
    \label{table:dataset}
\end{table}

\begin{table*}[t]
\centering
\renewcommand{\arraystretch}{1.2}
    \resizebox{\linewidth}{!}{%
    \begin{tabular}{l|l|cccccc}
    \toprule
   Method & Dataset & PGGAN & StarGAN & StyleGAN & StyleGAN2 & Bedroom & Bird \\
    \midrule
    \textbf{\sysname} (\textit{EfficientNet-B0})           
    & Bedroom & 86.25\% & 88.08\% & 90.47\% & 90.25\% & \underline{94.25\%} & 90.15\% \\
    & Bird & 87.80\% & 78.32\% & 98.49\% & 98.49\% & 97.63\% & \underline{88.82\%} \\
    \bottomrule
    \end{tabular}}
    \caption{Transfer learning results of non-face GAN-images. Dataset column indicates pre-trained model from source dataset, and Dataset row indicates test set of target dataset for transfer learning. Evaluation metric is AUROC (\%).}
\label{table:nonface}
\end{table*}

\begin{table*}[t]
\centering
\renewcommand{\arraystretch}{1.2}
    \resizebox{\linewidth}{!}{%
    \begin{tabular}{l|c|c|c|c|c} 
        \toprule
         Method & Base model & Source dataset & Target dataset & Source AUROC & Target AUROC \\
        \midrule
         w self-training & EfficientNet-B0 & StarGAN & PGGAN & \underline{\textbf{99.15}\%} & \textbf{94.94}\% \\
         w/o self-training & EfficientNet-B0 & StarGAN & PGGAN & \underline{98.96\%} & 92.54\% 
         \\
         \midrule
         w augmentation & EfficientNet-B0 & PGGAN & StyleGAN2 & \underline{\textbf{95.08}\%} & 98.13\% \\
         w/o augmentation & EfficientNet-B0 & PGGAN & StyleGAN2 & \underline{85.04\%} & 99.38\% \\
        \bottomrule
    \end{tabular}}
    \caption{Ablation study for self-training and data augmentation. The augmentation includes intra-class Cutmix, JPEG compression, Gaussian blur, and random horizontal flip. Our model with self-training shows a 2.40\% higher target AUROC than those without self-training, increasing the target AUROC from 92.54\% to 94.94\%. Our model with augmentation shows a 10.04\% higher source AUROC than those without augmentation. The underlined results are the source dataset performance after transfer learning. The best results are highlighted in bold.}
\label{table:perform_self}
\end{table*}

\section{Experimental Results}
The description of our dataset and training details are presented in Section D and E, respectively.

\subsection{Baselines}
\textbf{General transfer learning method.\hspace{1mm}} It is common practice to freeze some weights of pre-trained model from source dataset, and fine-tune the model with weight decay to the target dataset. We also experiment with them for GAN-image detection and call the method GeneralTransfer. We freeze all layers except for the top block layers and FC layers, and the base model is EfficientNet-B0, while GeneralTransfer is trained for 500 epochs with low learning (0.001) and momentum rates (0.1). Our method differs in that we train all weights and regularize them. The rest of the process is the same, e.g., noise injection to the model and the input data. 

\textbf{ForensicTransfer.\hspace{1mm}} We implement ForensicTransfer as the baseline in our experiment, where we trained it for 30 epochs in the pre-training stage and for 10 epochs in the transfer stage. The data usage is identical as in our method, but the difference appears in data augmentation and learning strategies. In our experiment, we follow the same data augmentation and learning strategy as ForensicTransfer.

\subsection{Performance Evaluation}
\textbf{AUROC metric.\hspace{1mm}} We use Area Under Receiver Operating characteristic Curve (AUROC) to evaluate the model. A broader area under the AUROC indicates a stronger model whose prediction is well classified by a decision boundary. In our work, the AUROC is more suitable for evaluating models than accuracy, since the AUROC does not require a threshold.


\textbf{Pre-trained model performance.\hspace{1mm}} We present our overall performance results in Table~\ref{table:performance}, where the same test datasets are used as in earlier section. In the zero-shot category, all diagonal terms of results (underlined results) represent single dataset performance of pre-trained models, as shown in the 3rd column of Table~\ref{table:performance}. The baselines and ~\sysname~have strong GAN-image detection ability, most of which achieving over 99\% AUROC.

\textbf{GeneralTransfer vs.~\sysname.\hspace{1mm}} GeneralTransfer freezes the pre-trained weights and fine-tunes the model, learning from the target dataset. In the Transfer Learning category, GeneralTransfer shows a trade-off between the source and target performance: after transfer learning, we observe high AUROC for the source dataset (99.86\% from pre-trained PGGAN), but low performance for transfer learning (54.17\%, and 54.18\% AUROC to StyleGAN and StyleGAN2, respectively), or forgetting of the learned source dataset (47.37\% AUROC from pre-trained StyleGAN2), but high AUROC for the target dataset (91.33\%, and 88.16\% for StyleGAN and StarGAN, respectively). On the contrary,~\sysname~(\textit{EfficientNet-B0}) shows consistent source dataset results: PGGAN (95.87\%), StarGAN (97.32\%), StyleGAN (97.85\%), and StyleGAN2 (97.71\%).~\sysname~is well transferred on the target dataset. In particular, it achieves 98.12\%, and 98.13\% AUROC for StyleGAN and StyleGAN2, respectively, maintaining 95.87\% AUROC from the pre-trained PGGAN.

\textbf{ForensicTransfer vs.~\sysname.\hspace{1mm}} ForensicTransfer shows some generalizability, but yields low performance; With pre-trained PGGAN, StyleGAN, and StyleGAN2, ForensicTransfer achieves 69.35\%, 65.83\%, and 65.84\%, respectively. It also clearly shows a trade-off between performance of the source and target datasets. Using pre-trained StarGAN for transfer learning on PGGAN, PGGAN performance shows 90.14\%, but forgets the learned information from StarGAN (51.32\%).

\textbf{ResNext vs. EfficientNet.\hspace{1mm}} In comparison to~\sysname~from different base models, the results show subtle differences. EfficientNet-B0 has generally stronger performance in GAN-image detection with fewer parameters. Although the number of parameters affects the classification performance, the performance of EfficientNet (3M) was superior to that of ResNext (20M) in transfer tasks. Therefore, T-GD performance is not directly related to the number of parameters.

\textbf{Non-face GAN-image detection.\hspace{1mm}}\sysname~is effective not only for GAN-generated face detection, but also for non-face tasks. We experimented with transfer learning from non-face GAN-images as the source (PGGAN-images from LSUN-bedroom and LSUN-bird) to face GAN-images as the target. We achieved stable AUROC on both detection tasks as shown in Table~\ref{table:nonface}.

\begin{figure}
\centering
    \includegraphics[width=0.50\textwidth]{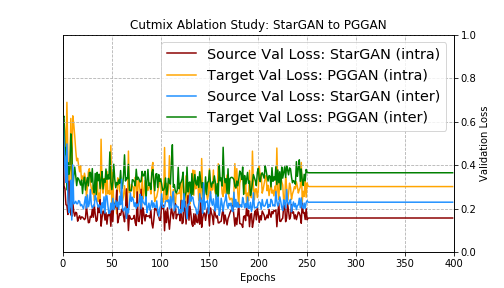}
    \caption{Validation loss in transfer learning between Cutmix and intra-class Cutmix. We can observe that after 50 epochs, the validation loss for intra-class Cutmix (yellow and red) is considerably lower and more stable than that for Cutmix (green and blue).}
    \label{fig:cutmix_analysis}
\end{figure}

\subsection{Ablation Study}
\textbf{Self-training effect.\hspace{1mm}} In Section 3.2, we explained why self-training is used for $L^2$-$SP$. In this ablation study, we validate this method through an ablation study. As shown in Table~\ref{table:perform_self}, we experiment with~\sysname, which is pre-trained on the StarGAN dataset and transferred to the PGGAN dataset, to compare the performance of our model with and without self-training, while keeping all other settings the same. For the target dataset with self-training, we achieved an AUROC that is 2.40\% higher than that of the target dataset without self-training (from 92.54\% to 94.94\%). The AUROC of the source dataset also increased from 98.96\% to 99.15\%, as shown in Table~\ref{table:perform_self}.

\textbf{Data augmentation effect.\hspace{1mm}} We utilized the following data augmentation methods to avoid over-fitting in transfer learning: intra-class Cutmix, JPEG compression, Gaussian blur, and random horizontal flip. We experiment with these augmentation methods through an ablation study. The performance of our model pre-trained on the PGGAN dataset and transferred to the StyleGAN2 dataset, with and without augmentation, is shown in Table~\ref{table:perform_self}. For the target dataset with augmentation, the AUROC of ~\sysname~dropped from 99.38\% to 98.13\% (1.25\%), but we achieved a 10.04\% higher AUROC for the source dataset than that of the same dataset without augmentation (from 85.04\% to 95.08\%). Despite the small reduction in the target AUROC, the drastic increase in the source AUROC implies that over-fitting can be avoided through these augmentation methods in transfer learning, while preventing \textit{catastrophic forgetting}.

\textbf{Inter-Cutmix vs. Intra-class Cutmix.\hspace{1mm}} In Fig.~\ref{fig:cutmix_analysis}, the X-axis and the Y-axis represent the training epochs and the validation loss, respectively. The validation loss performance of Cutmix is shown on green and blue, while that of intra-class Cutmix is presented on yellow and red. Red and blue lines represent the respective validation loss for the source dataset (StarGAN) and, yellow and green represent the target dataset (PGGAN) with respect to epochs. The model pre-trained on the StarGAN dataset is transferred to the PGGAN dataset. The base model is EfficientNet-B0, and all other settings remain the same except for Cutmix. Our experiment shows that intra-class Cutmix has a lower validation loss, and thus a higher performance for both the source and target datasets. We conclude that the improved performance of intra-class Cutmix attributes to the fact that GAN-image detection is a binary classification problem, where the two classes are real and GAN-generated. Mixing only two labels can be harmful and degrade the classification performance.

\textbf{Grad-CAM.\hspace{1mm}} We perform a qualitative analysis of each GAN-image output by using Gradient Class Activation Map (Grad-CAM). The results are presented in Supp. Section C. We describe the results obtained from Grad-CAM, which visualizes the essential regions from an input image required for the prediction of its class.



\section{Conclusion}
We present T-GD network, a method to maintain high performance on both the source and target datasets for the GAN-image detection during transfer learning. We propose the novel regularization and augmentation techniques, the $L^2$-$SP$ self-training and intra-class Cutmix, building upon well-known CNN backbone models. While previous research focused on leveraging the metadata information from different GAN models, our method outperforms over other approaches on both source and target datasets without using any metadata from the GAN models. 
In particular, when PGGAN-images are used as the source data for transfer learning, we observe the best transfer learning performance. Therefore, we recommend PGGAN as the guided dataset for the source data. As the GAN detection classifier evolves, the new generation methods will also appear in the future. Hence, the lack of training data from new GAN generators will be a significant problem. To cope with this issue, we plan to work on classifying GAN-images with few-shot or zero-shot learning. We also hope that future work will continue to challenge and improve existing transfer learning strategies.
Our code is available here.\footnote{\url{https://github.com/cutz-j/T-GD}}
\section*{Acknowledgements}
We thank Siho Han for providing his expertise to greatly improve this work. This work was partly supported by Institute of Information \& communications Technology Planning \& Evaluation (IITP) grant funded by the Korea government (MSIT) (No.2019-0-00421, AI Graduate School Support Program (Sungkyunkwan University)), and the National Research Foundation of Korea (NRF) grant funded by the Korea government (MSIT) (No. 2017R1C1B5076474, and 2020R1C1C1006004). Also, this research was results of a study on the ``HPC Support'' Project, supported by the `Ministry of Science and ICT' and NIPA. Additionally, this work was partly supported by Institute for Information \& communication Technology Promotion (IITP) grant funded by the Korea government (MSIT) (No. 2019-0-01343, Regional strategic industry convergence security core talent training business).

\bibliography{bibtex}

\begin{thebibliography}{36}
\providecommand{\natexlab}[1]{#1}
\providecommand{\url}[1]{\texttt{#1}}
\expandafter\ifx\csname urlstyle\endcsname\relax
  \providecommand{\doi}[1]{doi: #1}\else
  \providecommand{\doi}{doi: \begingroup \urlstyle{rm}\Url}\fi

\bibitem[Choi et~al.(2019)Choi, Uh, Yoo, and Ha]{choi2019stargan}
Choi, Y., Uh, Y., Yoo, J., and Ha, J.-W.
\newblock Stargan v2: Diverse image synthesis for multiple domains.
\newblock \emph{arXiv preprint arXiv:1912.01865}, 2019.

\bibitem[Chollet(2017)]{chollet2017xception}
Chollet, F.
\newblock Xception: Deep learning with depthwise separable convolutions.
\newblock In \emph{Proceedings of the IEEE conference on computer vision and
  pattern recognition}, pp.\  1251--1258, 2017.

\bibitem[Cozzolino et~al.(2018)Cozzolino, Thies, R{\"o}ssler, Riess,
  Nie{\ss}ner, and Verdoliva]{cozzolino2018forensictransfer}
Cozzolino, D., Thies, J., R{\"o}ssler, A., Riess, C., Nie{\ss}ner, M., and
  Verdoliva, L.
\newblock Forensictransfer: Weakly-supervised domain adaptation for forgery
  detection.
\newblock \emph{arXiv preprint arXiv:1812.02510}, 2018.

\bibitem[Huang et~al.(2016)Huang, Sun, Liu, Sedra, and
  Weinberger]{huang2016deep}
Huang, G., Sun, Y., Liu, Z., Sedra, D., and Weinberger, K.~Q.
\newblock Deep networks with stochastic depth.
\newblock In \emph{European conference on computer vision}, pp.\  646--661.
  Springer, 2016.

\bibitem[Ioffe \& Szegedy(2015)Ioffe and Szegedy]{ioffe2015batch}
Ioffe, S. and Szegedy, C.
\newblock Batch normalization: Accelerating deep network training by reducing
  internal covariate shift.
\newblock \emph{arXiv preprint arXiv:1502.03167}, 2015.

\bibitem[Jeon et~al.(2019)Jeon, Bang, and Woo]{jeon2019faketalkerdetect}
Jeon, H., Bang, Y., and Woo, S.~S.
\newblock Faketalkerdetect: Effective and practical realistic neural talking
  head detection with a highly unbalanced dataset.
\newblock In \emph{Proceedings of the IEEE International Conference on Computer
  Vision Workshops}, pp.\  0--0, 2019.

\bibitem[Jeon et~al.(2020)Jeon, Bang, and Woo]{jeon2020fdftnet}
Jeon, H., Bang, Y., and Woo, S.~S.
\newblock Fdftnet: Facing off fake images using fake detection fine-tuning
  network.
\newblock \emph{arXiv preprint arXiv:2001.01265}, 2020.

\bibitem[Karras et~al.(2017)Karras, Aila, Laine, and
  Lehtinen]{karras2017progressive}
Karras, T., Aila, T., Laine, S., and Lehtinen, J.
\newblock Progressive growing of gans for improved quality, stability, and
  variation.
\newblock \emph{arXiv preprint arXiv:1710.10196}, 2017.

\bibitem[Li et~al.(2020)Li, Chaudhari, Yang, Lam, Ravichandran, Bhotika, and
  Soatto]{li2020rethinking}
Li, H., Chaudhari, P., Yang, H., Lam, M., Ravichandran, A., Bhotika, R., and
  Soatto, S.
\newblock Rethinking the hyperparameters for fine-tuning.
\newblock \emph{arXiv preprint arXiv:2002.11770}, 2020.

\bibitem[Li et~al.(2018)Li, Grandvalet, and Davoine]{li2018explicit}
Li, X., Grandvalet, Y., and Davoine, F.
\newblock Explicit inductive bias for transfer learning with convolutional
  networks.
\newblock \emph{arXiv preprint arXiv:1802.01483}, 2018.

\bibitem[Li et~al.(2019)Li, Xiong, Wang, Rao, Liu, and Huan]{li2019delta}
Li, X., Xiong, H., Wang, H., Rao, Y., Liu, L., and Huan, J.
\newblock Delta: Deep learning transfer using feature map with attention for
  convolutional networks.
\newblock \emph{arXiv preprint arXiv:1901.09229}, 2019.

\bibitem[Marra et~al.(2019{\natexlab{a}})Marra, Gragnaniello, Verdoliva, and
  Poggi]{marra2019gans}
Marra, F., Gragnaniello, D., Verdoliva, L., and Poggi, G.
\newblock Do gans leave artificial fingerprints?
\newblock In \emph{2019 IEEE Conference on Multimedia Information Processing
  and Retrieval (MIPR)}, pp.\  506--511. IEEE, 2019{\natexlab{a}}.

\bibitem[Marra et~al.(2019{\natexlab{b}})Marra, Saltori, Boato, and
  Verdoliva]{marra2019incremental}
Marra, F., Saltori, C., Boato, G., and Verdoliva, L.
\newblock Incremental learning for the detection and classification of
  gan-generated images.
\newblock \emph{arXiv preprint arXiv:1910.01568}, 2019{\natexlab{b}}.

\bibitem[Nataraj et~al.(2019)Nataraj, Mohammed, Manjunath, Chandrasekaran,
  Flenner, Bappy, and Roy-Chowdhury]{nataraj2019detecting}
Nataraj, L., Mohammed, T.~M., Manjunath, B., Chandrasekaran, S., Flenner, A.,
  Bappy, J.~H., and Roy-Chowdhury, A.~K.
\newblock Detecting gan generated fake images using co-occurrence matrices.
\newblock \emph{Electronic Imaging}, 2019\penalty0 (5):\penalty0 532--1, 2019.

\bibitem[Qiao et~al.(2019)Qiao, Wang, Liu, Shen, and Yuille]{qiao2019weight}
Qiao, S., Wang, H., Liu, C., Shen, W., and Yuille, A.
\newblock Weight standardization.
\newblock \emph{arXiv preprint arXiv:1903.10520}, 2019.

\bibitem[RoyChowdhury et~al.(2019)RoyChowdhury, Chakrabarty, Singh, Jin, Jiang,
  Cao, and Learned-Miller]{roychowdhury2019automatic}
RoyChowdhury, A., Chakrabarty, P., Singh, A., Jin, S., Jiang, H., Cao, L., and
  Learned-Miller, E.
\newblock Automatic adaptation of object detectors to new domains using
  self-training.
\newblock In \emph{Proceedings of the IEEE Conference on Computer Vision and
  Pattern Recognition}, pp.\  780--790, 2019.

\bibitem[Russakovsky et~al.(2015)Russakovsky, Deng, Su, Krause, Satheesh, Ma,
  Huang, Karpathy, Khosla, Bernstein, Berg, and Fei-Fei]{ILSVRC15}
Russakovsky, O., Deng, J., Su, H., Krause, J., Satheesh, S., Ma, S., Huang, Z.,
  Karpathy, A., Khosla, A., Bernstein, M., Berg, A.~C., and Fei-Fei, L.
\newblock {ImageNet Large Scale Visual Recognition Challenge}.
\newblock \emph{International Journal of Computer Vision (IJCV)}, 115\penalty0
  (3):\penalty0 211--252, 2015.
\newblock \doi{10.1007/s11263-015-0816-y}.

\bibitem[Shaham et~al.(2019)Shaham, Dekel, and Michaeli]{shaham2019singan}
Shaham, T.~R., Dekel, T., and Michaeli, T.
\newblock Singan: Learning a generative model from a single natural image.
\newblock In \emph{Proceedings of the IEEE International Conference on Computer
  Vision}, pp.\  4570--4580, 2019.

\bibitem[Srivastava et~al.(2014)Srivastava, Hinton, Krizhevsky, Sutskever, and
  Salakhutdinov]{srivastava2014dropout}
Srivastava, N., Hinton, G., Krizhevsky, A., Sutskever, I., and Salakhutdinov,
  R.
\newblock Dropout: a simple way to prevent neural networks from overfitting.
\newblock \emph{The journal of machine learning research}, 15\penalty0
  (1):\penalty0 1929--1958, 2014.

\bibitem[Szegedy et~al.(2017)Szegedy, Ioffe, Vanhoucke, and
  Alemi]{szegedy2017inception}
Szegedy, C., Ioffe, S., Vanhoucke, V., and Alemi, A.~A.
\newblock Inception-v4, inception-resnet and the impact of residual connections
  on learning.
\newblock In \emph{Thirty-first AAAI conference on artificial intelligence},
  2017.

\bibitem[Tan \& Le(2019)Tan and Le]{tan2019efficientnet}
Tan, M. and Le, Q.~V.
\newblock Efficientnet: Rethinking model scaling for convolutional neural
  networks.
\newblock \emph{arXiv preprint arXiv:1905.11946}, 2019.

\bibitem[Tariq et~al.(2018)Tariq, Lee, Kim, Shin, and Woo]{tariq2018detecting}
Tariq, S., Lee, S., Kim, H., Shin, Y., and Woo, S.~S.
\newblock Detecting both machine and human created fake face images in the
  wild.
\newblock In \emph{Proceedings of the 2nd international workshop on multimedia
  privacy and security}, pp.\  81--87, 2018.

\bibitem[Tariq et~al.(2019)Tariq, Lee, Kim, Shin, and Woo]{tariq2019gan}
Tariq, S., Lee, S., Kim, H., Shin, Y., and Woo, S.~S.
\newblock Gan is a friend or foe? a framework to detect various fake face
  images.
\newblock In \emph{Proceedings of the 34th ACM/SIGAPP Symposium on Applied
  Computing}, pp.\  1296--1303, 2019.

\bibitem[Wang et~al.(2019)Wang, Wang, Zhang, Owens, and Efros]{wang2019cnn}
Wang, S.-Y., Wang, O., Zhang, R., Owens, A., and Efros, A.~A.
\newblock Cnn-generated images are surprisingly easy to spot... for now.
\newblock \emph{arXiv preprint arXiv:1912.11035}, 2019.

\bibitem[Wu \& He(2018)Wu and He]{wu2018group}
Wu, Y. and He, K.
\newblock Group normalization.
\newblock In \emph{Proceedings of the European Conference on Computer Vision
  (ECCV)}, pp.\  3--19, 2018.

\bibitem[Xie et~al.(2019)Xie, Hovy, Luong, and Le]{xie2019self}
Xie, Q., Hovy, E., Luong, M.-T., and Le, Q.~V.
\newblock Self-training with noisy student improves imagenet classification.
\newblock \emph{arXiv preprint arXiv:1911.04252}, 2019.

\bibitem[Xie et~al.(2017)Xie, Girshick, Doll{\'a}r, Tu, and
  He]{xie2017aggregated}
Xie, S., Girshick, R., Doll{\'a}r, P., Tu, Z., and He, K.
\newblock Aggregated residual transformations for deep neural networks.
\newblock In \emph{Proceedings of the IEEE conference on computer vision and
  pattern recognition}, pp.\  1492--1500, 2017.

\bibitem[Xuan et~al.(2019)Xuan, Peng, Wang, and Dong]{xuan2019generalization}
Xuan, X., Peng, B., Wang, W., and Dong, J.
\newblock On the generalization of gan image forensics.
\newblock In \emph{Chinese Conference on Biometric Recognition}, pp.\
  134--141. Springer, 2019.

\bibitem[Yalniz et~al.(2019)Yalniz, J{\'e}gou, Chen, Paluri, and
  Mahajan]{yalniz2019billion}
Yalniz, I.~Z., J{\'e}gou, H., Chen, K., Paluri, M., and Mahajan, D.
\newblock Billion-scale semi-supervised learning for image classification.
\newblock \emph{arXiv preprint arXiv:1905.00546}, 2019.

\bibitem[Yim et~al.(2017)Yim, Joo, Bae, and Kim]{yim2017gift}
Yim, J., Joo, D., Bae, J., and Kim, J.
\newblock A gift from knowledge distillation: Fast optimization, network
  minimization and transfer learning.
\newblock In \emph{Proceedings of the IEEE Conference on Computer Vision and
  Pattern Recognition}, pp.\  4133--4141, 2017.

\bibitem[Yu et~al.(2019)Yu, Davis, and Fritz]{yu2019attributing}
Yu, N., Davis, L.~S., and Fritz, M.
\newblock Attributing fake images to gans: Learning and analyzing gan
  fingerprints.
\newblock In \emph{Proceedings of the IEEE International Conference on Computer
  Vision}, pp.\  7556--7566, 2019.

\bibitem[Yun et~al.(2019)Yun, Han, Oh, Chun, Choe, and Yoo]{yun2019cutmix}
Yun, S., Han, D., Oh, S.~J., Chun, S., Choe, J., and Yoo, Y.
\newblock Cutmix: Regularization strategy to train strong classifiers with
  localizable features.
\newblock In \emph{Proceedings of the IEEE International Conference on Computer
  Vision}, pp.\  6023--6032, 2019.

\bibitem[Zakharov et~al.(2019)Zakharov, Shysheya, Burkov, and
  Lempitsky]{zakharov2019few}
Zakharov, E., Shysheya, A., Burkov, E., and Lempitsky, V.
\newblock Few-shot adversarial learning of realistic neural talking head
  models.
\newblock In \emph{Proceedings of the IEEE International Conference on Computer
  Vision}, pp.\  9459--9468, 2019.

\bibitem[Zamir et~al.(2018)Zamir, Sax, Shen, Guibas, Malik, and
  Savarese]{zamir2018taskonomy}
Zamir, A.~R., Sax, A., Shen, W., Guibas, L.~J., Malik, J., and Savarese, S.
\newblock Taskonomy: Disentangling task transfer learning.
\newblock In \emph{Proceedings of the IEEE Conference on Computer Vision and
  Pattern Recognition}, pp.\  3712--3722, 2018.

\bibitem[Zenke et~al.(2017)Zenke, Poole, and Ganguli]{zenke2017continual}
Zenke, F., Poole, B., and Ganguli, S.
\newblock Continual learning through synaptic intelligence.
\newblock \emph{Proceedings of machine learning research}, 70:\penalty0 3987,
  2017.

\bibitem[Zhang et~al.(2019)Zhang, Karaman, and Chang]{zhang2019detecting}
Zhang, X., Karaman, S., and Chang, S.-F.
\newblock Detecting and simulating artifacts in gan fake images.
\newblock \emph{arXiv preprint arXiv:1907.06515}, 2019.

\end{thebibliography}


\begin{thebibliography}{4}
\providecommand{\natexlab}[1]{#1}
\providecommand{\url}[1]{\texttt{#1}}
\expandafter\ifx\csname urlstyle\endcsname\relax
  \providecommand{\doi}[1]{doi: #1}\else
  \providecommand{\doi}{doi: \begingroup \urlstyle{rm}\Url}\fi

\bibitem[Choi et~al.(2018)Choi, Choi, Kim, Ha, Kim, and Choo]{choi2018stargan}
Choi, Y., Choi, M., Kim, M., Ha, J.-W., Kim, S., and Choo, J.
\newblock Stargan: Unified generative adversarial networks for multi-domain
  image-to-image translation.
\newblock In \emph{Proceedings of the IEEE conference on computer vision and
  pattern recognition}, pp.\  8789--8797, 2018.

\bibitem[Karras et~al.(2019{\natexlab{a}})Karras, Laine, and
  Aila]{karras2019style}
Karras, T., Laine, S., and Aila, T.
\newblock A style-based generator architecture for generative adversarial
  networks.
\newblock In \emph{Proceedings of the IEEE Conference on Computer Vision and
  Pattern Recognition}, pp.\  4401--4410, 2019{\natexlab{a}}.

\bibitem[Karras et~al.(2019{\natexlab{b}})Karras, Laine, Aittala, Hellsten,
  Lehtinen, and Aila]{karras2019analyzing}
Karras, T., Laine, S., Aittala, M., Hellsten, J., Lehtinen, J., and Aila, T.
\newblock Analyzing and improving the image quality of stylegan.
\newblock \emph{arXiv preprint arXiv:1912.04958}, 2019{\natexlab{b}}.

\bibitem[Liu et~al.(2015)Liu, Luo, Wang, and Tang]{liu2015deep}
Liu, Z., Luo, P., Wang, X., and Tang, X.
\newblock Deep learning face attributes in the wild.
\newblock In \emph{Proceedings of the IEEE international conference on computer
  vision}, pp.\  3730--3738, 2015.

\end{thebibliography}
\bibliographystyle{icml2020}

\begin{appendices}
\clearpage
\onecolumn
\nocitesec{*}
\section{Pipeline of Self-training for \bm{$L^2$}-\bm{$SP$}}

\begin{algorithm}[h]
  \caption{Self-training for $L^2$-$SP$}
  \label{alg:selftraining}
\begin{algorithmic}
    \REQUIRE a source data $\{(x_1, y_1),(x_2,y_2),...,(x_n,y_n)\}$ and a target data $\{(\tilde{x}_1,\tilde{y}_1),(\tilde{x}_2, \tilde{y}_2),...,(\tilde{x}_m,\tilde{y}_m)\}$.
    \STATE{\bfseries 1:} Pre-train teacher model $w^\prime$ and minimizes the cross-entropy loss on the source dataset.
    \begin{ALC@g}
        \STATE {\bfseries Input:} Data ($x_i, y_i$), size $n$, output $\hat{y}_i$
        \STATE {\bfseries Objective:} $\min_w\sum^{n}_{i=1}J(\hat{y_i}, y_i)+\lambda\cdot\Omega_{l2}(w)$ \\ 
    \end{ALC@g}
    \STATE{\bfseries 2:} $w^\prime$ is used as a starting point for learning the student model.
    \begin{ALC@g}
        \STATE {\bfseries Input:} Data ($\tilde{x}_i^{noised}, \tilde{y}_i^{noised}$), size $m$, output $\hat{y}_{i}^{\prime}$
        \STATE {\bfseries $\gamma$ score:} $\sigma(-\frac{1}{n}\sum_{i=1}^{n}J(\hat{y}_{i}^{\prime},\tilde{y}_i))$
    \end{ALC@g}
    \STATE{\bfseries 3:} Learn student model $w$ and minimizes the cross-entropy loss and the regularization terms with $\gamma$.
    \begin{ALC@g}
        \STATE {\bfseries Input:} Data ($\tilde{x}_i^{noised}, \tilde{y}_i^{noised}$), size $m$, output $\hat{y}_{i}^{\star}$
        \STATE {\bfseries Objective:} $\min_w\sum^{n}_{i=1}J(\hat{y}_i^{\star},y_i)+\gamma\cdot\Omega_{sp}(w,w^\prime)+\gamma\cdot\Omega_{l2}(w_{fc})$
    \end{ALC@g}
    \STATE{\bfseries 4:} Feedback: Use the student model as the teacher model and go back to step {\bfseries 2}
\end{algorithmic}
\end{algorithm}

\textbf{Pipeline.\hspace{1mm}} Figure~\ref{fig:selftraining} depicts the overall self-training process, and Algorithm~\ref{alg:selftraining} describes the detailed process. The inputs are labeled data from the target dataset. We separated the learning of the teacher model (pre-training classifiers on the source dataset) from that of the student model (fine-tuning the transferred model). The algorithm shows a self-training method, in which the teacher and the student learn by exchanging feedback. Feedback refers to the process in which we copy the weight of the student onto that of the teacher at a pre-defined cycle, which we chose to be 200. When the student model is trained (the stage of transfer learning), augmented input data and the noise-injected model are used. It renders the transfer learning more robust during the training process.

\begin{figure}[h!]
    \begin{center}
      \includegraphics[width=0.5\linewidth]{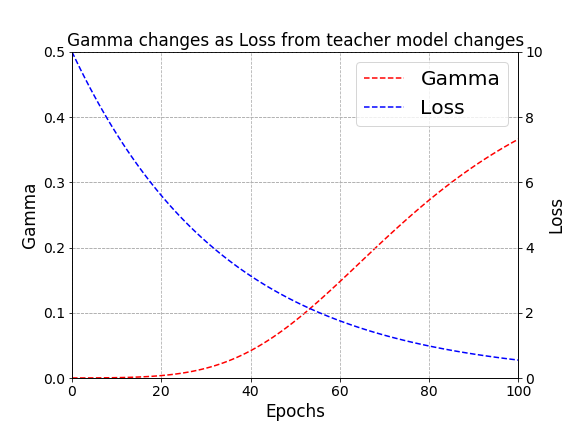}
    \end{center}
       \caption{Self-training parameter $\gamma$ variation. Each red line and blue line represents cross-entropy loss $J(\hat{y_i},\tilde{y}_i)$ from the teacher model and $\gamma$ which varies accordingly. The left-Y-axis, right-Y-axis, and X-axis represent Gamma value, right-Y-axis, loss $J(\hat{y_i},\tilde{y}_i)$}
    \label{fig:sigmoid}
\end{figure}

\textbf{Gamma.\hspace{1mm}} Figure~\ref{fig:sigmoid} shows how $\gamma$ changes as $J(\hat{y_i},\tilde{y}_i)$ changes. $\gamma$ depends on the loss of a large dataset. For a large loss of the target dataset, the target model requires more training to achieve smaller loss and cause $\gamma$ to reduce. On the other hand, for a small loss, learning from the target dataset requires less training and cause $\gamma$ to increase. In an early training phase, the teacher loss is very high, because the model was not trained. As a result, $\gamma$ is close to zero, and the regularization of the $L2$-$SP$ and $L^2$-norm is weakened. As the target is trained, the loss decreases, while $\gamma$ increases, meaning that the regularization is stronger; figure~\ref{fig:sigmoid} indicates this variation. This allows us to avoid an extreme regularization effect, either excessive or insufficient, which is a critical issue when using the fixed hyperparameter.
T-GD avoids the error amplification for two reasons: first, we used a small volume of labeled data, unlike the typical self-training method using a large volume of unlabeled data, and second, $\gamma$ is scaled within a range of 0 to 0.5 by the sigmoid function. Consequently, extremely high/low loss values 
of the teacher due to false information do not have a significant impact on the learning.

\section{Intra-class Cutmix}

\setlength{\intextsep}{1\baselineskip}
\begin{algorithm}[h]
    \caption{Intra-class Cutmix algorithm}
    \begin{algorithmic}[1]
        \STATE\algorithmicrequire{a target data $\{(x_1, y_1),(x_2,y_2),...,(x_n,y_n)\}$}, where the input data is composed of tensors of size $m\times c\times w\times h$, where $m$ is the mini-batch size, $c$ is the channel (3), $w$ is the width (128), and $h$ is the height (128).
        \FOR {each epoch}
            \STATE$X_m$, $Y_m$ = $\{(x_i,y_i),...,(x_m,y_m)\}$
            \IF {training}
                \STATE$X^\prime_m$, $Y^\prime_m$ = random\_shuffle($X_m$, $Y_m$)
                \STATE same\_index = [$Y^\prime_m$==$Y_m$]
                \STATE$\rho$ = Uniform(0, 1)
                \IF {$\rho \geq$ cutmix\_prob}
                    \STATE $b_x$, $b_y$ = Uniform(0, $w$), Uniform(0, $h$)
                    \STATE $b_w$, $b_h$ = Sqrt(1 - $\lambda$), Sqrt(1 - $\lambda$)
                    \STATE $x1$, $x2$ = Round(Clip($b_x - b_w / 2$, min=0)), Round(Clip($b_x + b_w / 2$, max=$w$))
                    \STATE $y1$, $y2$ = Round(Clip($b_y - b_h / 2$, min=0)), Round(Clip($b_y + b_h / 2$, max=$h$))
                    \STATE$X_m$[same\_index,:,$x1$:$x2$,$y1$:$y2$] = $X^\prime_m$[same\_index,:,$x1$:$x2$,$y1$:$y2$]
                \ENDIF
            \ENDIF
            \STATE $\hat{Y}_m$ = feed\_forward($X_m$)
            \STATE $J$ = loss\_function($\hat{Y}_m$, $Y_m$)
            \STATE update()
        \ENDFOR
    \end{algorithmic}
\label{alg:cutmix}
\end{algorithm}

\begin{figure}[h!]
\centering
    \includegraphics[width=0.9\textwidth]{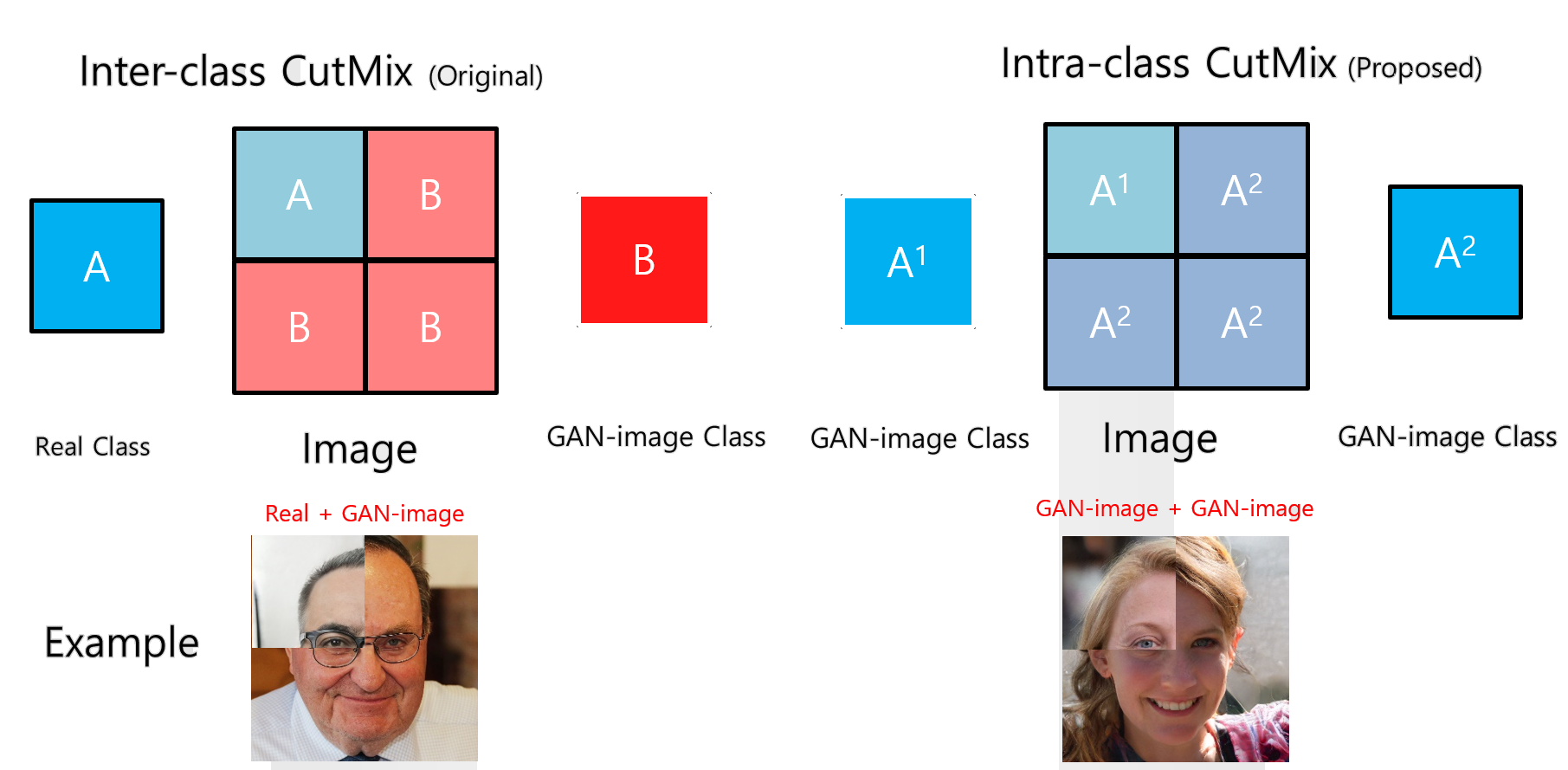}
    \caption{Comparison between Cutmix and proposed Intra-class Cutmix.}
    \label{fig:icm}
\end{figure}

Algorithm~\ref{alg:cutmix} presents the pseudo-code for the intra-class Cutmix. Figure~\ref{fig:icm} describes the difference between Cutmix and Intra-class Cutmix. On the left side, the original CutMix, is referred to as Inter-class CutMix. Inter-class Cutmix replaces the chosen patch with another image patch in the same location. The ground truth labels are also mixed proportionally to the area of the patches. On the right side is shown our proposed Intra-class CutMix. The ground truth is not used in the mixing region. In our experiment, we found that the inter-class CutMix for a binary classification causes highly unstable training.

\newpage
\section{Gradient Class Activation Map (Grad-CAM)}
\begin{figure}[h!]
\centering
    \includegraphics[width=0.9\textwidth]{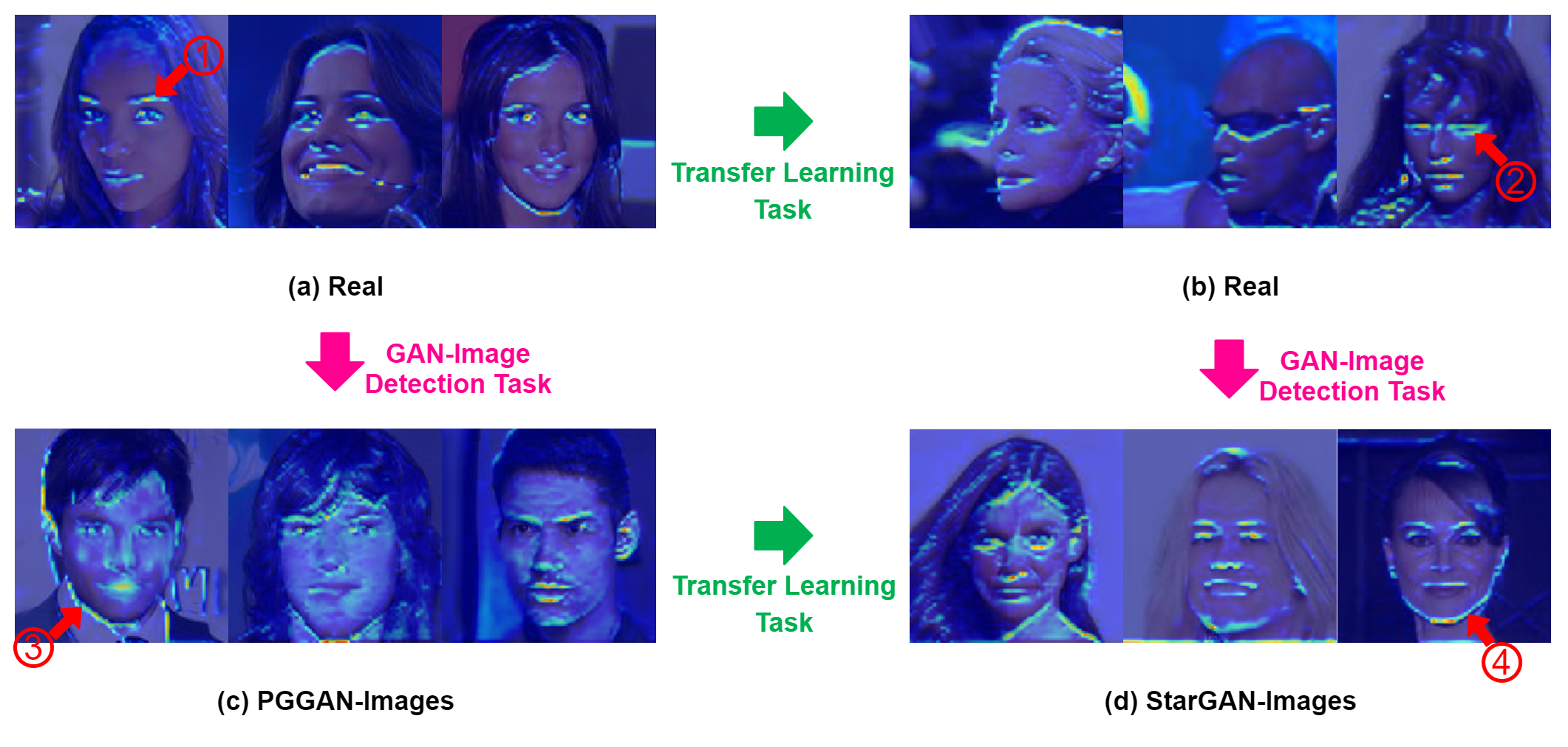}
    \caption{Class activation map output of the images. (a) and (b) show the CelebA-HQ real images for PGGAN and StarGAN, respectively; (c) and (d) show the GAN-images from PGGAN and StarGAN, respectively. In the images tinged with blue, we can observe that activated facial regions are highlighted. Moreover, different levels of intensity are represented via varying highlight colors, where regions highlighted in yellow show the strongest activation, while those highlighted in green and sky blue show weaker activations. Red circled numbers from 1 to 4 point to edge representations, such as eyebrows, glasses, and the jawline.}
    \label{fig:Grad-CAM}
\end{figure}

In this section, we describe the results obtained from Grad-CAM, which visualizes the essential regions from an input image required for the prediction of its class.

\textbf{Grad-CAM.\hspace{1mm}} Grad-CAM is generated based on the gradient between the input image and the predicted class. Using this heat map, we can measure how the layer output affects the prediction by evaluating the pixel values: positive pixels resulting from the convolution and ReLU layers translate to activated regions in Grad-CAM represented by fluorescent color, while negative pixels show no activation in blue color. For this experiment, we utilized EfficientNet-B0 as the base model, pre-trained on the PGGAN dataset, and then transferred to the StarGAN dataset. 

\begin{figure}[b!]
\centering
    \includegraphics[width=0.4\textwidth]{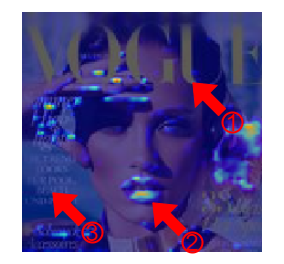}
    \caption{Grad-CAM output of a real image with sharp representations. Although both characterized as edge representations, the lips are activated (circled number 2), while the word "VOGUE" in the foreground and the words in the background are not (circled numbers 1 and 3).}
    \label{fig:pg_real_sharp}
\end{figure}

\textbf{Edge representation in~\sysname.\hspace{1mm}} Figure~\ref{fig:Grad-CAM} (a) and (b) show the Grad-CAMs for the real class, and (c) and (d) show those of the GAN-image class. For the task of GAN-image detection (magenta arrow in Fig.~\ref{fig:Grad-CAM}), we observe that both Grad-CAMs focus on the edges of the face, indicated by circled numbers. For the task of transfer learning (green arrow in Fig.~\ref{fig:Grad-CAM}), we observe that similar regions of the face are highlighted, showing distinct activations on the jawline for both the PGGAN and StarGAN datasets as indicated by the circled numbers 3 and 4 in Fig.~\ref{fig:Grad-CAM}. This implies that the pre-trained model on the source dataset has been successfully transferred to the target dataset.

\textbf{Facial edge representation with sharp representations.\hspace{1mm}} Our experiment shows that~\sysname~can effectively distinguish facial representations from letters present in the foreground and the background, as indicated by the circled numbers 1 and 3 in Fig.~\ref{fig:pg_real_sharp}, respectively.

\newpage
\section{Dataset Description}
We describe three real and four GAN-generated image dataset we used in our experiment.  

\textbf{CelebA.\hspace{1mm}} 
CelebFaces Attributes Dataset (CelebA)~\cite{liu2015deep} is a large-scale face attributes dataset with more than 200,000 celebrity images.

\textbf{CelebA-HQ.\hspace{1mm}}
CelebA-High-quality (CelebA-HQ) consists of 30,000 images~\cite{liu2015deep}. This applied various image processing to center the images on the facial region. 

\textbf{FFHQ.\hspace{1mm}}
Provided by StyleGAN~\cite{karras2019style}, Flickr-Faces-HQ (FFHQ) consists of 70,000 high-quality images crawled from Flickr at a 1024$\times$1024 resolution. The images represent individuals of different ages and ethnicities, contain various backgrounds, and have much better coverage of accessories, such as eyeglasses, sunglasses, and hats, compared to CelebA-HQ.

\textbf{PGGAN.\hspace{1mm}} 
The key idea of PGGAN~\cite{karras2017progressive} is to grow the generator and the discriminator progressively. Model training starts at a low resolution, with the addition of layers to increase the spatial resolution of the generated images. For PGGAN-images, we used the official implementation dataset\footnote{\url{https://github.com/tkarras/progressive_growing_of_gans}} provided by the author, consisting of 100,000 GAN-generated fake celebrity images at a 1024$\times$1024 resolution generated from the CelebA-HQ dataset. For our experiment, we resized each image to a 128$\times$128 resolution.

\textbf{StarGAN.\hspace{1mm}}
StarGAN~\cite{choi2018stargan} is capable of learning mappings among multiple domains using only a single model that can  generate image-to-image translated high quality images. For the image generation, we used the official implementation source code and CelebA dataset~\cite{liu2015deep} to generated 128$\times$128 resolution GAN-images. We generated StarGAN-images from this model and insure that we follow their official implementation by using their pre-trined model\footnote{\url{https://github.com/yunjey/stargan}}. We generated five attributes GAN-images from one CelebA image: black-hair, blond-hair, brown-hair, male, and young attributes. Then, we randomly chose one of five images as the source dataset. 

\textbf{StyleGAN.\hspace{1mm}}
StyleGAN~\cite{karras2019style} architecture leads to an automatically learned, unsupervised separation of high-level attributes and stochastic variation in the generated images, enabling an intuitive and scale-specific control of the synthesis process.
For StyleGAN-images, we used the official implementation dataset\footnote{\url{https://github.com/NVlabs/stylegan}} provided by the author, consisting of 100,000 GAN-generated celebrity images at a 1024$\times$1024 resolution generated from the FFHQ~\cite{karras2019style} dataset. For our experiment, we resized the image to a 256$\times$256 resolution.

\textbf{StyleGAN2.\hspace{1mm}}
StyleGAN2~\cite{karras2019analyzing} redesigns the generator normalization, revisits the progressive growing, and regularizes the generator to encourage a good conditioning when mapping latent vectors to images. For StyleGAN2-images, we used the official implementation dataset\footnote{\url{https://github.com/NVlabs/stylegan2}} provided by the author, under the same condition as in StyleGAN~\cite{karras2019style}.

\section{Training Details}
We implement EfficientNet-B0~\cite{tan2019efficientnet} and ResNext32$\times$4d~\cite{xie2017aggregated}. We change BN to GN and WS for better transferability. For both pre-training teacher models, we use a batch size of 512, stochastic gradient descent (SGD) optimizer with a momentum 0.9, and gradual warm-up start by 4 times for 20 epochs with cosine-annealing. The initial learning rate is 0.04 and epochs are 300. Different data augmentation techniques are applied: JPEG compression (0.2 rate), Gaussian Blur (0.2), intra-class Cutmix (0.2), random horizontal flip (0.2), dropout (0.2), and stochastic depth (0.2). In the stage of transfer learning, we use a batch size of 200, SGD optimizer with a momentum 0.1, and an initial learning rate of 0.01. All augmentation rates are set to 0.5, except for dropout and stochastic depth: JEPG compression (0.5), Gaussian Blur (0.5), intra-class Cutmix (0.5), random horizontal flip (0.5), dropout (0.2), and stochastic depth (0.2). The training is completed at 1000 iterations.
\bibliographysec{appendix_bibtex}
\bibliographystylesec{icml2020}

\end{appendices}
\end{document}